\documentclass[10pt,letterpaper]{article}
\usepackage[top=0.85in,left=2.75in,footskip=0.75in]{geometry}

\usepackage{amsmath,amssymb}

\usepackage{changepage}

\usepackage[utf8x]{inputenc}

\usepackage{textcomp,marvosym}

\usepackage{cite}
\usepackage[hyphens]{url}

\usepackage{nameref,hyperref}

\usepackage[right]{lineno}

\usepackage{microtype}
\DisableLigatures[f]{encoding = *, family = * }

\usepackage[table]{xcolor}

\usepackage{enumitem}

\usepackage{array}

\usepackage{multirow}

\usepackage[absolute]{textpos}

\usepackage{ragged2e}

\newcolumntype{+}{!{\vrule width 2pt}}

\newlength\savedwidth

\raggedright
\usepackage[aboveskip=1pt,labelfont=bf,labelsep=period,justification=raggedright,singlelinecheck=off]{caption}

\makeatletter
\renewcommand{\@biblabel}[1]{\quad#1.}
\makeatother

\usepackage{lastpage,fancyhdr,graphicx}
\usepackage{epstopdf}
\PassOptionsToPackage{hyphens}{url}\usepackage{hyperref}

\fancyheadoffset[L]{2.25in}
\fancyfootoffset[L]{2.25in}
\begin{document}
\vspace*{0.2in}

\begin{textblock}{3.5}(1,2)
\textblocklabel{block two}
\noindent
\justify
\scriptsize We present this Open Access paper as a service to interested readers. 

\noindent
The original paper appeared as: Tsotsos JK, Kotseruba I, Wloka C (2019) Rapid visual categorization is not guided by early salience-based selection. PLoS ONE 14(10): e0224306.\  \url{https://doi.org/10.1371/journal.pone.0224306}\\pmid:31648265 but  had 3 incorrect figures.

\noindent
This version replaces those figures (noted in Tsotsos JK, Kotseruba I, Wloka C (2019) Correction: Rapid visual categorization is not guided by early salience-based selection. PLOS ONE 14(12): e0226429 \ \url{ https://doi.org/10.1371/journal.pone.0226429}), and includes the full supplementary material for the  convenience of a single, complete pdf.

\end{textblock}

\begin{textblock}{3.5}(1,6)
\textblocklabel{block two}
\noindent
\scriptsize \textbf{Citation:} Tsotsos JK, Kotseruba I, Wloka C (2019) Rapid visual categorization is not guided by early salience-based selection. PLoS ONE 14(10): e0224306. \url{https://doi.org/10.1371/journal. pone.0224306}
\par
\bigskip
\noindent
\scriptsize\textbf{Editor:} Mudassar Raza, COMSATS University Islamabad, Wah Campus, PAKISTAN
\par
\bigskip
\noindent
\textbf{Received:} February 20, 2019
\par
\bigskip
\noindent
\textbf{Accepted:} October 11, 2019
\par
\bigskip
\noindent
\textbf{Published:} October 24, 2019
\par
\bigskip
\noindent
Copyright: © 2019 Tsotsos et al. This is an open access article distributed under the terms of the Creative Commons Attribution License, which permits unrestricted use, distribution, and reproduction in any medium, provided the original author and source are credited.
\par
\bigskip
\noindent
\scriptsize\textbf{Data Availability Statement:} Two image datasets were used in this paper. One image dataset is from Thorpe et al, originally purchased from the Corel Stock Photo Library \url{https://www.amazon.com/Corel-Stock-Photo-Library-2/dp/B000V933GI}. The ground truth masks for this data created by the authors have been made publicly available at the URL below. 

 The other image dataset is from Potter et al. The authors confirm they have permission to share the Potter et al image data and ground truth masks created by the authors. These data are available from the Center for Open Science (\url{https://osf.io/cemkw/}). Additionally, all algorithms used in the study have publicly available code with URLs provided in the Supporting Information.

\end{textblock}

\begin{flushleft}
{\Large
\textbf\newline{Rapid Visual Categorization is not Guided by Early Salience-Based Selection} 
}
\newline
\\
John K. Tsotsos\textsuperscript{1}*,
Iuliia Kotseruba\textsuperscript{1},
Calden Wloka\textsuperscript{1}
\\
\bigskip
\textsuperscript{1}Department of Electrical Engineering and Computer Science, York University, Toronto, ON, Canada
\\
\bigskip

%
%





*Corresponding author\\
E-Mail: tsotsos@eecs.yorku.ca (JKT)

\end{flushleft}
\section*{Abstract}
The current dominant visual processing paradigm in both human and machine research is the feedforward, layered hierarchy of neural-like processing elements. Within this paradigm, visual saliency is seen by many to have a specific role, namely that of early selection. Early selection is thought to enable very fast visual performance by limiting processing to only the most salient candidate portions of an image. This strategy has led to a plethora of saliency algorithms that have indeed improved processing time efficiency in machine algorithms, which in turn have strengthened the suggestion that human vision also employs a similar early selection strategy. However, at least one set of critical tests of this idea has never been performed with respect to the role of early selection in human vision. How would the best of the current saliency models perform on the stimuli used by experimentalists who first provided evidence for this visual processing paradigm? Would the algorithms really provide correct candidate sub-images to enable fast categorization on those same images? Do humans really need this early selection for their impressive performance? Here, we report on a new series of tests of these questions whose results suggest that it is quite unlikely that such an early selection process has any role in human rapid visual categorization.



\section*{Introduction}
The current dominant visual processing paradigm in both human and machine research is the learned, feedforward, layered hierarchical network of neural-like processing elements, with a history stretching from Rosenblatt's Perceptrons \cite{rosenblatt1965principles}, Fukushima's Cognitron \cite{fukushima1975cognitron} (and subsequent Neocognitron \cite{fukushima1982neocognitron}), to Rumelhart \& McClelland's Parallel Distributed Processes \cite{rumelhart1986parallel}, LeCun \& Bengio's Convolutional Neural Networks \cite{lecun1995convolutional}, and to Krizhevsky, Sutskever, and Hinton's Deep Neural Networks \cite{krizhevsky2012imagenet}. The goal of each of these models and systems was to explain or emulate the effortless ability of humans to immediately perceive content in images. Tsotsos \cite{tsotsos1988complexity} termed this \textit{immediate vision} and laid out the computational difficulty of the task as well as key elements of how brains and machines might defeat its combinatorial nature.

Our everyday experience tells us that vision feels immediate: we simply open our eyes and the world is there, fully formed before us and ready for our interactions. There is no perceptible time delay or inner `turning of wheels'. It is well-documented over several decades that humans can recognize visual targets with remarkably short exposure times, with the seminal works including Potter \& Levy\cite{potter1969recognition}, Potter \& Faulconer \cite{potter1975time}, Potter \cite{potter1975meaning}, Thorpe et al. \cite{thorpe1996speed}, and more recently Potter et al. \cite{potter2014detecting}. The short exposure times (the shortest being 13ms) and subsequent fast responses (150ms of neural processing required for yes-no answers to categorize an image) led theoreticians to conclude that there was no time available for any processing other than a single pass through the visual hierarchy in the feedforward direction \cite{feldman1982connectionist}. 

\begin{textblock}{3.5}(1,1.3)
\textblocklabel{block two}
\noindent
\scriptsize\textbf{Funding:} This research was supported by grants to the senior author (JKT) from the following sources: Air Force Office of Scientific Research USA (FA9550-18-1-0054) (\url{https://www.wpafb.af.mil/afrl/afosr/}), Office of Naval Research USA (N00178-16-P-0087) (\url{https://www.onr.navy.mil/}), The Canada Research Chairs Program (950- 231659) (\url{http://www.chairs-chaires.gc.ca/home-accueil-eng.aspx}) and Natural Sciences and Engineering Research Council of Canada (RGPIN-2016-05352) (\url{https://www.canada.ca/en/science-engineering-research.html}). The funders had no role in study design, data collection and analysis, decision to publish, or preparation of the manuscript.
\justify
\noindent
\scriptsize \textbf{Competing interests:} The authors have declared that no competing interests exist.
\end{textblock}

To be sure, there are a variety of models and theories that add feedback and recurrence to such hierarchical networks from Fukushima \cite{fukushima1986neural} to
Hochreiter \& Schmidhuber \cite{hochreiter1997long} and Sutskever \cite{sutskever2012training} and more, but with respect to the main thrust of this paper, these do not detract from the main conclusions here because they address tasks different from rapid visual categorization and thus, generally, would be inconsistent with the observed time course of human categorization performance.

Still, the computational requirements to fully process a whole scene seem daunting \cite{tsotsos1989complexity} and many suggested that there must be some sort of optimizing action to reduce computational load occurring along that feedforward path. Within this paradigm, the processing of visual saliency has been suggested to have this specific role, namely reducing computational load via early selection \cite{koch1987shifts}. Early selection was thought to reduce the information that must be processed to enable very fast visual performance by determining a spatial region-of-interest (ROI) on which further analysis should be performed. In the Koch \& Ullman formulation \cite{koch1987shifts}, a saliency map is computed early in the visual processing stream and represents point-wise stimulus conspicuity (contrast between a point and its local surround). A winner-take-all competition selects the most conspicuous location (point) and the features at the selected location are routed to a central representation for further processing. It is important to note that Koch and Ullman viewed saliency as a method for predicting the most useful image locations for processing in recognition or similar higher level tasks; eye movements were not considered as an outcome of saliency computation in their paper. Inhibition of that selected location forces a shift to the next most conspicuous location when the algorithm is run again. Koch \& Ullman's early selection idea seems to have been motivated by Feature Integration Theory \cite{treisman1980feature} in that it provided a mechanistic version of its operation, specifically, the selection of a focus within the master map of locations. It shares much with the Broadbent's Early Selection model, which was based on human auditory behavior \cite{broadbent1958perception}. In the first stage `physical' properties (e.g. pitch) would be extracted for all incoming (auditory) stimuli, in a `parallel' manner and in the second stage, psychological properties, beyond simple physical characteristics (e.g. meaning of spoken words) would be extracted. This second stage was more limited in capacity, so that it could not deal with all the incoming information at once when there were multiple stimuli (having to process them ‘serially’, rather than in parallel). A selective filter protected the second stage from overload, passing to it only those stimuli which had a particular physical property, from among those already extracted for all stimuli within the first stage. Many criticized Broadbent's early selection idea and proposed alternates including late selection schemes \cite{deutsch1963attention}, \cite{mackay1973aspects}, \cite{moray1969attention}, \cite{norman1968toward}, and attenuation schemes \cite{treisman1964effect}.

Early implementations of saliency computation did indeed produce points of maximum conspicuity that were found useful for machine vision  \cite{clark1988modal}, \cite{sandon1990simulating}, \cite{culhane1992attentional}, \cite{itti1998model}. As algorithms evolved, they moved more towards salient region or object proposals (for reviews see \cite{bylinskii2015towards}, \cite{bruce2015computational}, \cite{bylinskii2018different}). Fixation-based algorithms are typically validated by how well they match human fixation points (even though eye movements were not part of the original experimental work nor are they the only manifestation of attentional behaviour), while salient object detection algorithms  are validated by how well the regions they produce overlap with ground truth object outlines or bounding boxes. The number of different saliency conceptualizations and implementations now is in the hundreds \cite{tsotsos2015computational}. Models based on deep learning methods have also recently embraced this early selection idea in the hopes that their already impressive success can be improved further \cite{ba2014multiple}, \cite{zhang2018top}.

Many high-profile models of human visual information processing, appearing over the past 3 decades, include some variant of early selection within a feedforward visual processing stream \cite{Shashua1988StructuralST}, \cite{olshausen1993neurobiological}, \cite{itti2001computational}, \cite{walther2002attentional}, \cite{li2002saliency}, \cite{zhaoping2014understanding}, \cite{deco2004neurodynamical}, \cite{itti2005models}, \cite{chikkerur2010and}, \cite{zhang2011object}, \cite{buschman2015behavior}. These not only claim biological inspiration but also biological realism. That is, the authors claim that their processing strategies actually reflect the brain's visual processing strategy. Often, it is difficult to evaluate such claims. For example, Yan et al. \cite{yan2018bottom} claim that their
evidence regarding V1 representation during an orientation singleton
task, where a monkey learns to make an eye movement to a singleton
oriented bar, reflects bottom-up salience and go as far as to assert that V1 computes a bottom-up saliency map. Their experiments do not address whether any of the other representations throughout the
visual cortex have a similar characteristic. Whether or not such representation exists in the brain has been addressed by many, all of whom find supporting evidence for a saliency map, including: superior colliculus \cite{horwitz1999separate}, \cite{kustov1996shared} \cite{mcpeek2002saccade}; LGN \cite{koch1984theoretical}, \cite{sherman1986control}; pulvinar \cite{petersen1987contributions}, \cite{posner1990attention}, \cite{robinson_petersen_1992}; FEF \cite{thompson1997dissociation}; parietal areas \cite{gottlieb1998representation}. In each of these, the connection to a saliency representation is made because maxima of responses that are found within a neural population correspond with the behaviorally attended location. The Yan et al. work also draws their conclusion based on this observation. Could it be that they all do simultaneously?

It almost seems a straightforward inference that if V1 shows this characteristic, each visual area receiving feedforward input from V1 necessarily also shows it, and this continues through the visual hierarchy. The use of machine learning methods to demonstrate ‘read out’ of neural response only shows that it is possible to extract the necessary information from a neural population and not that this is the actual sole source of that information. Since the information that would lead to behavior is necessarily in the stimulus itself, any representation of that stimulus that is created in a sufficiently non-destructive manner necessarily also includes that same information. In making claims about a single locus for a saliency map, these authors fail to provide a mechanistic explanation for how behavior is generated directly from that representation and without any relevant influence from other brain processing areas. To demonstrate the existence of a saliency map in any representation, one must present evidence that some retinotopic visual area alone has a causal connection to behavior. It is also important to recall that eye movements, the behavior linked to saliency in Yan et al. \cite{yan2018bottom}, played no role in the original conceptualization of the saliency map.

\section*{Hypothesis}

The present work examines one of the basic underlying features of all these models: that salience-based early selection within a feedforward visual hierarchy provides a spatial ROI on which further analysis is performed (i.e., the basic Koch \& Ullman idea described above). Our motivation was born from the observation that at least one set of critical tests of this idea has never been performed. No one has tested saliency models on the stimuli that were used in the seminal experiments that supported the feedforward view. Do these algorithms really provide an accurate reduction of the visual search space to enable fast categorization? Here, the first question we ask is: If the computation of saliency occurs early in the feedforward pass through visual areas, and determines a location for further processing, does the first ROI determined by a saliency algorithm effectively point to the target?

If the question were to be answered in the affirmative, then when the images used in the seminal experiments are run through a saliency algorithm, the algorithm should yield a prediction for the target location that matches well with the ground truth. It would be reasonable to assume that a good prediction is followed by a correct categorization because a good prediction identifies the target sufficiently well. That only the first selection of an algorithm is of interest is key: the temporal constraints provided by the experimental observations and which theorists used  (e.g., Feldman \& Ballard \cite{feldman1982connectionist}) do not permit more than one selection. 

After conducting this experiment, the results pointed us to a second question: is human rapid visual categorization guided by early selection? This led to a second experiment, this time examining human behaviour, where we manipulated the stimulus image set to discover the target location and extent needed for good categorization. While addressing these two questions, we also examined other issues, specifically, center bias in datasets and algorithm biological plausibility. Together, the natural conclusion is that early selection played no role in the key seminal experiments. 

\section*{Methodology}
There are several main elements that comprise our approach to the questions raised about early selection: the image data sets, the set of algorithms tested, the performance metrics used, the analysis of algorithm biological plausibility, and human experiments (approved by the York University Office of Research Ethics, certificate 2016-014 ``Selective tuning approach to visual system attention executive'') to examine performance to parafoveal stimuli from the image data sets. Each will be described in turn.

Before these descriptions, the overall logic of this argument and the role of these components is presented. If early selection via image saliency plays a role in rapid human visual categorization, then we hoped to find existing saliency algorithms, that generally have very strong performance on available benchmarks, that approached human performance in their ability to predict targets. If we succeeded, then the good algorithms would point to potential directions for how they might be  improved and utilized for machine vision. We also wanted to answer the early selection question for human vision so needed to examine the algorithms not only for their accuracy but also whether they embodied basic biological constraints known to underlie human categorization behavior. We could thus say whether the human design inspiration was useful. When we realized that we were finding mostly negative results on these counts, we wondered whether humans needed any such early guidance at all and tested this by creating a full set of parafoveal stimuli for the same categorization task. The results on all of these experiments point to the facts that the development of salience algorithms is not yet at a point where human level performance can be expected and that humans might have no need of such early guidance for the original categorization tasks of the seminal experiments in any case. In order to answer all of the above questions we developed a set of metrics with which we compared algorithm and human performance.

\subsection*{Image sets and summary of original results}
The following seminal experiments and image datasets were considered. Potter \& Levy \cite{potter1969recognition} examined memory for visual events occurring at and near the rate of eye fixations. Their subjects were shown sequences of 16 pictures, from 272 magazine photos, presented with rates of 0.5, 1, 2, 3, 4, 6, or 8 images per second. They concluded that rapidly presented pictures are processed separately for precisely the time each is in view and are not held with other items in a short-term memory. This was among the earliest works to demonstrate rapid categorization but the image set was unavailable for our use.

Potter and Faulconer \cite{potter1975time} used 96 stimuli, half of these being line drawings and half words that represent objects in those line drawings. There were 18 categories of objects and within each, between 2 and 9 instances (e.g., food: carrot, pie; clothing: hat, coat; tools: pliers, hammer). Each stimulus was preceded and followed by a mask of random lines and pieces of letters. Target information was provided to subjects in some trials before and, in others, after the stimulus. Stimuli were shown for 40, 50, 60 or 70ms. They observed that subjects needed 44ms exposure duration for the drawings and 46ms for the words to achieve 50\% accuracy. We obtained these images and tested some of the saliency algorithms; however, we did not pursue these. The saliency algorithms all produced simply blurred versions of the line drawings and thus were not useful, likely due to their development being primarily based on natural images. This description is included here because it sets an early data point for fast categorization.

Potter \cite{potter1975meaning} used an RSVP (Rapid Serial Visual Presentation) task with 16 photos of natural scenes. Subjects were either shown the target that they might find within the sequence in advance or were told its name in advance. Accuracy was over 70\% after only 125ms of exposure. In a second experiment, she tested subject's memory. Subjects looked at a 16-image sequence of pictures without prior instruction and were then asked a yes-no question about what they had seen. Subjects required about 300ms exposure to achieve 50\% accuracy. Regrettably, this dataset was also unavailable.

\begin{figure}[!h]
\includegraphics[width=\linewidth]{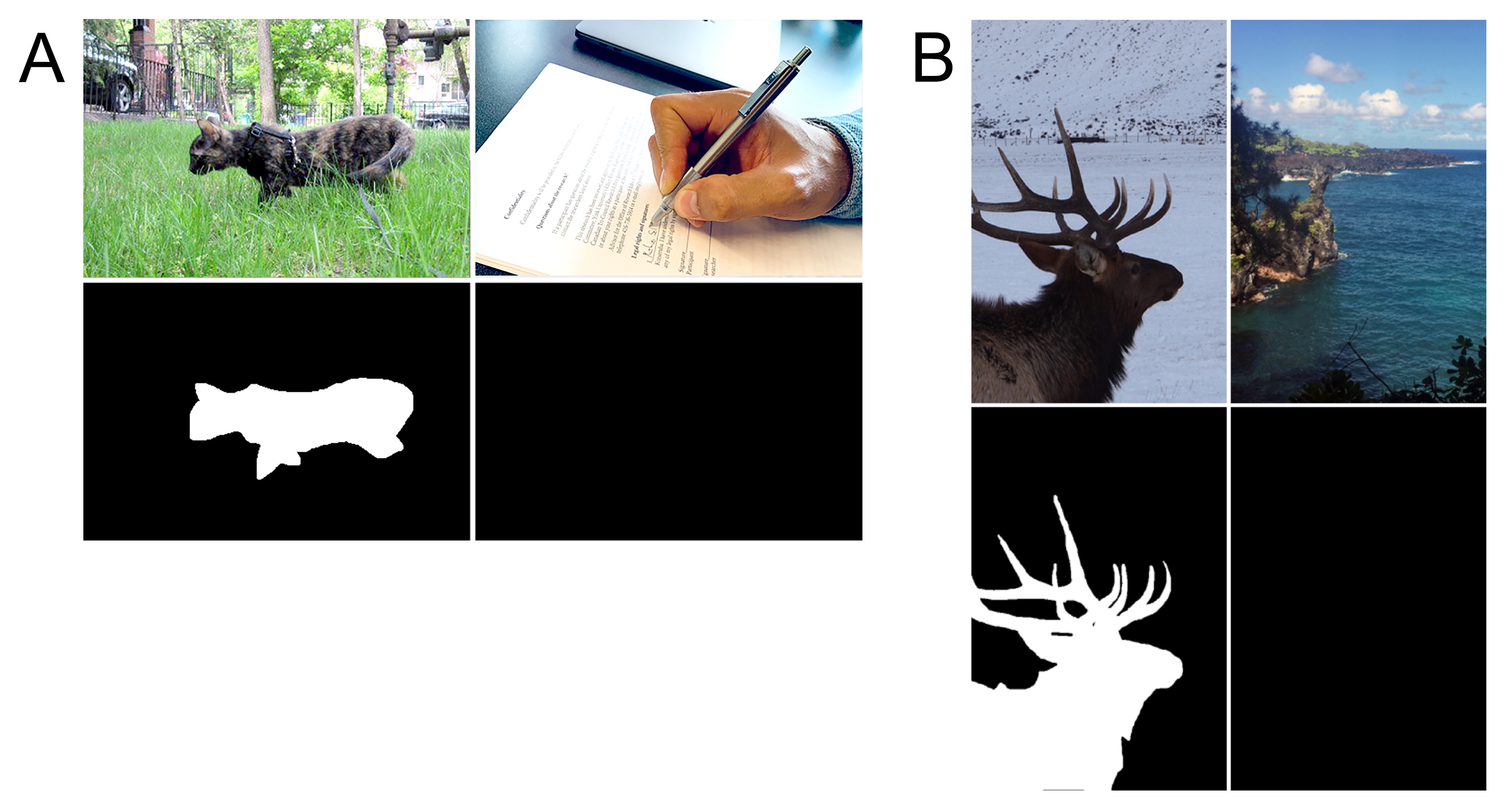}
\caption{{\bf Sample categorization images.} A: Sample images representative of those from the Potter dataset with target present (top left), no target present (top right), and corresponding ground truth masks (bottom row). In the experiment, participants were asked if they saw a particular target (here, ``a walking cat''). Only those trials where the question was asked after the stimulus presentation were used. The bottom panels show the hand-drawn ground truth masks; no-target images have no ground-truth. B: Sample images, representative of those in the Thorpe
dataset with target present (top left), no target present (top right), and corresponding ground truth masks (bottom row). In the experiment participants were asked if there is an animal in the image. The bottom panels show the hand-drawn
ground truth masks; no-target images have no ground-truth.}
\label{fig1}
\end{figure}

Fortunately, there was an alternate stimulus set that was available. Potter et al. \cite{potter2014detecting} used an RSVP task of a series of six or 12 color pictures presented at 13, 27, 53 and 80ms per picture, with no inter-stimulus interval. Images were 300x200 in size, and there were 1711 images in total, 366 with target present. An example is shown in Fig~\ref{fig1}A, with the corresponding hand-drawn ground truth mask (GTM). Participants were to detect the presence or absence of a target specified by a name (e.g., smiling couple) that was given just before or immediately after the sequence (in other words, subjects only had usable target expectations in half the trials). If subjects reported a positive detection, they were then asked  a 2-alternative forced choice (2AFC) question to see if they could recognize the target given a distractor. Detection improved with increasing duration (from 13ms up to 80ms) and was generally better when the name was presented before the sequence, but performance was significantly above chance at all durations, whether the target was named before or after the sequence. At the shortest exposure, prior knowledge seemed to provide no benefit at all. For the set of trials without prior expectations, the ones relevant to our study, performance of the 2AFC task ranged from about 67\% to about 73\% correct on the target-present trials. Performance when prior expectation was provided ranged from 75-85\% accuracy. The results are consistent with feedforward models, in which an initial wave of neural activity through the ventral stream is sufficient to allow identification of a complex visual stimulus in a single forward pass. Potter and her colleagues generously provided this dataset for our work and confirmed that these stimuli were of the same type as used in \cite{potter1969recognition} and \cite{potter1975meaning}. 

Thorpe et al. \cite{thorpe1996speed} ran `yes-no' categorization tasks. Subjects viewed color images (with none repeated during trials), and had to determine if an animal was present or not. The original images were 512x768 in size but downsized in Thorpe’s experiment to 256x384 (which we used), totaled 2000 images, with 996 having target present. A representative example is shown in Fig~\ref{fig1}B. There was no prior knowledge of types of animals and stimuli were taken from commercial images. They measured behavior plus ERP (Event Related Potential). Even though the duration of the image exposure was 20ms, subjects exhibited 94\% average correctness. Prefrontal ERP activity diverged at 150ms after stimuli onset for ‘yes’ and ‘no’ responses, which means enough processing had been done in 150ms to decide if an animal is present or not. They concluded that sufficient processing must be occurring in a primarily feedforward manner. Thorpe and colleagues graciously provided the full original dataset for our research.

The two chosen experiments for our comparison are not identical and some justification as to why they are suitable for our test of feedforward saliency is in order.  In the Thorpe et al. case, the processing path is direct and very much what we need to compare against; if feedforward saliency computation is part of human categorization performance it would definitely be part of the 150ms time period Thorpe et al. reported. The most direct other experiment for us to include would have been the Potter 1975 paper\cite{potter1975meaning}. The closest we have to this is the Potter et al. stimulus set \cite{potter2014detecting}, confirmed to involve stimuli of the same type as the earlier paper. The detection component, which would reflect the same direct path as Thorpe et al., is present but was followed with an additional task. This means we should not compare time courses - and we do not. The detection task is reported using d’ values in the main paper but also using percent correct in their supplementary material, which is what we used. Only the results for trials where there was no prior knowledge are relevant for our work. They show that for target-present trials, proportion correct improved as stimulus duration increased from 13 ms to 80ms, from about 60\% to 73\% (we use 73\% as the performance mark), while for target-absent trials the mean correct was 75\%. These values were for their 6 picture test; for their 12-picture test, the accuracy was similar. We stress that our tests do not impact the validity of any of the original experiments cited.

\subsection*{The tested algorithm set}

To conduct our test, we chose 7 bottom-up fixation based saliency models. Each is referred to by the acronym in bold given here. Two are algorithms that represent a cross-section of classical methods: the most commonly used and cited model by Itti et al. (ITTI) \cite{itti1998model} and the AIM model \cite{bruce2009saliency}, a consistently high performing model in benchmark fixation tests. We also selected several recent algorithms which achieved high scores in the MIT benchmark \cite{bylinskii2015saliency} and had publicly available source code (see \nameref{S1_Text} for details), namely Saliency in Context (oSALICON - the open source version) \cite{huang2015salicon}, Boolean Map based Saliency (BMS) \cite{zhang2013saliency}, Ensembles of Deep Networks (eDN) by \cite{vig2014large}, RARE2012  \cite{riche2013rare2012}, and DeepGaze II \cite{kummerer2016deepgaze}. eDN, oSALICON, and DeepGaze II represent the class of saliency algorithms based on deep learning. eDN model is a set of shallow neuromophic networks selected via hyperparameter optimization for best performance on the MIT1003 saliency dataset \cite{judd2009learning}. The other two models rely on transfer learning from deep networks initially trained on object classification tasks (VGG-19 \cite{simonyan2014very} in DeepGaze II and VGG-16 \cite{simonyan2014very} in oSALICON) to achieve state-of-the-art performance on the MIT saliency benchmark. Finally, we also added the `objectness' algorithm (OBJ) because the human experiments all involve categorization of objects \cite{alexe2010object}. All algorithms were used with default parameters and published implementations. Many algorithms use an explicit center bias typically expressed as a centered Gaussian distribution in order to improve performance (typically a gain of 2-3\%). For those models the bias was disabled to enable a fair comarison. See \nameref{S1_Text} for implementation details. A 9th method was added for control purposes, which we refer to as CENTER. This places the point of interest at the center of the image regardless of image contents. We use $\mathcal{P}$ to denote the point of interest for all algorithms.

How one measures performance is very important especially when it involves direct comparison of human and machine output. In the absence of eye movements, the degree of overlap between a target and the region of high acuity in the retina is likely strongly correlated with human performance. As a result, some of our performance measures include this overlap. This seems easily justifiable by considering the details of human photoreceptor layout on the retina (see \nameref{S2_Text} for details). Observers in the original experiments were instructed to fixate the image center and there was no time for any change of gaze. Thus, if a sufficient spatial extent of the target lies within the observer's parafovea, it would seem that detection should be more likely correct. 

\begin{figure}[!h]
\includegraphics[width=\linewidth]{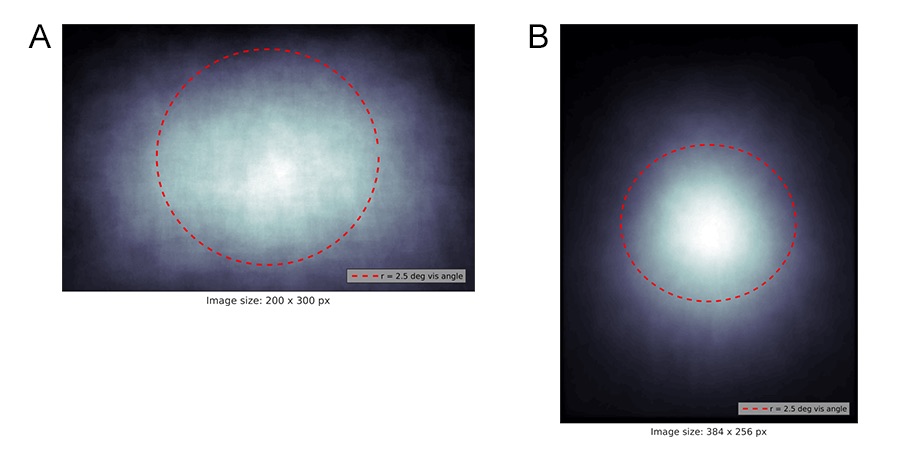}
\caption{{\bf Target distribution in datasets.} Two plots showing overlaid ground truth mask centroids for targets in the Potter (A) and Thorpe (B) datasets. Brighter pixels correspond to greater overlap between the masks. Parafovea (r=$2.5\,^{\circ}$) is shown by a dashed red line. Mean of the distribution lies approximately in the center of the image. The area within $2.5\,^{\circ}$ from the center of the image is calculated based on the following assumptions: the viewer is 57 cm away from the monitor, the monitor has 23" diagonal and resolution of 1920x1080.}
\label{fig2}
\end{figure}

If this assertion is appropriate, then image sets whose targets are strongly center-biased would lead to better categorization performance than those image sets with lesser bias. We thus created scatterplots for the target centroids of the two stimulus sets. This revealed a substantial center bias for the Potter set (Fig~\ref{fig2}A) and a strong center bias for the Thorpe set (Fig~\ref{fig2}B). Sure enough, Thorpe et al. reported higher human performance than Potter et al.

\subsection*{Performance metrics}

\begin{figure}[!h]
\includegraphics[width=\linewidth]{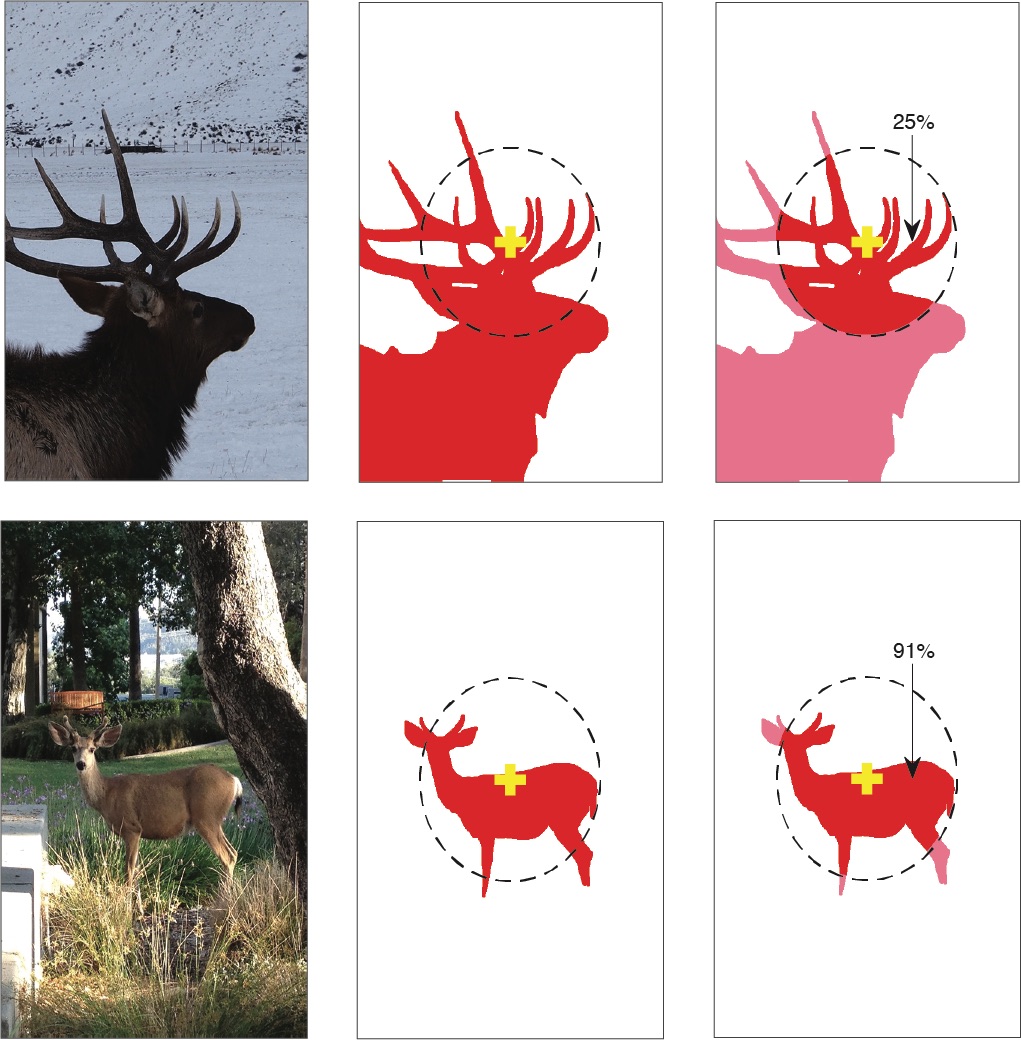}
\caption{{\bf Visualization of performance measures.} Two examples are shown. The original stimulus image is on the left with the subject's fixation point marked. The middle
image gives the ground truth mask where if the saliency algorithm produces a predicted $\mathcal{P}$, our measure A will count a positive hit. The right image shows the extent of the human parafovea, the dashed circle, centered at subject's fixation and superimposed on the GTM. Measure B will count a prediction as a hit if it falls within the marked area. Measure C and D will count a prediction as hit if measure B is a hit and the percentage of GTM area within the parafovea is sufficiently large. For these examples, 25\% of the elk GTM area lies within the parafovea while the deer is 91\% within. The elk image would lead to a `hit' only for measures A and B whereas all the measures would count the deer as a hit.}
\label{fig3}
\end{figure}

Our purpose was to determine guidelines for how we measure algorithm correctness during our tests. We considered several different ways of developing these metrics and a brief description of their derivation follows based on the human performance levels presented in the previous section. Thorpe et al. observed a 94\% accuracy in their experiment; 94\% of the targets had a least 27\% of their extent within the parafovea. Similarly, Potter et al. observed a peak accuracy of 73\% and 73\% of targets had at least 41\% of their area within the parafovea. To be sure, there is no correspondence between the set of observed correct responses, either per subject nor collectively, and the set of ground truth masks identified with this analysis. But such a correspondence cannot be computed with the available data and our purpose was not to correctly determine this correspondence. Samples of this calculation are shown in Fig~\ref{fig3} (the example with the image of an elk (top row) demonstrates where this assumption may be inappropriate).

Using these estimates, which are admittedly coarse at best, we limited the image region where a saliency algorithm prediction would be considered as a valid ROI cue for human categorization. Note that this is not a measure of algorithm correctness in the manner usually used in benchmark tests \cite{borji2015salient}. The goal is to quantify how well saliency algorithms provide guidance for the human visual system towards the task of accurate image categorization. In any case, these are only two of the performance measures; the other two have no similar approximate nature.

We thus decided on four separate ways of quantifying algorithm performance. A saliency algorithm's predicted first point of interest, $\mathcal{P}$, would be marked as correct if:

\begin{enumerate}[label=\Alph*)]
	\item{$\mathcal{P}$ is anywhere within the GTM;}
	\item{$\mathcal{P}$ is within the GTM AND within the parafovea, anywhere (even if by one pixel);}
	\item{$\mathcal{P}$ falls within both the GTM AND the parafovea AND at least 27\% of the GTM (by area) lies within the parafovea. This reflects the reality of Thorpe et al. data and will be applied only for those stimuli;}
	\item{$\mathcal{P}$ falls within the GTM AND within the parafovea AND at least 41\% of the GTM (by area) lies within the parafovea. This reflects the reality of Potter et al. data and will be applied only for those stimuli.}
\end{enumerate}

When compared to the observed target layout characteristics, these are conditions which are very generous in favor of the algorithms. One additional point must be addressed. None of the tested algorithms include the capacity to accept prior instruction. The Potter et al. results we use as comparison are those without subjects receiving prior guidance while those of Thorpe included uniform guidance for expected category. Although this might appear to lead to an unfair comparison, we note that the performance in the Potter et al.  experiment without prior expectation was roughly 10\% lower than with prior guidance \cite{potter2014detecting}. As a result, we can reasonably assume a similar decrease in the Thorpe experiment, and as will be seen, this will not affect the overall conclusion.

\begin{figure}[!h]
\includegraphics[width=\linewidth]{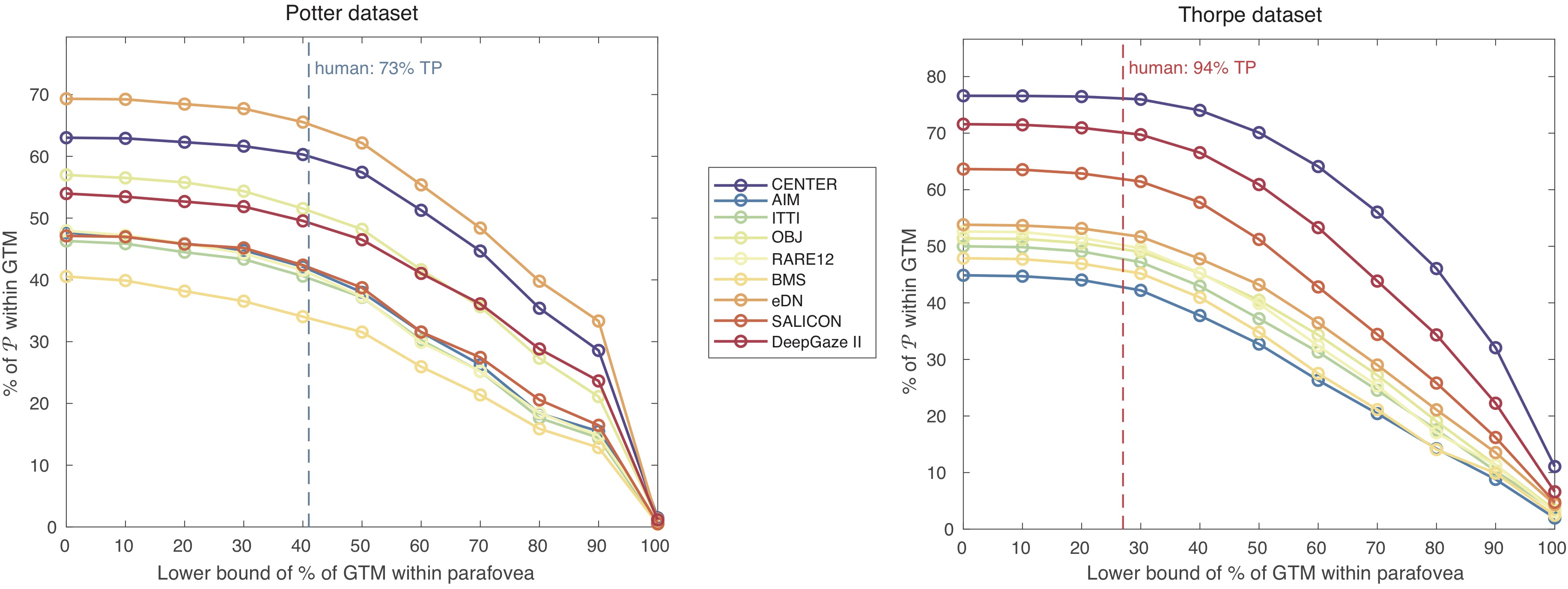}
\caption{{\bf Sensitivity Analysis of Performance Thresholds.} The plots show how percentage of points of interest P within the ground truth masks and parafovea gradually decreases depending on how much of the GTM (by area) is inside  the parafovea. On each plot the point (0,0) corresponds to measure B (i.e. P is within the GTM and parafovea regardless of the amount of GTM within the parafovea). Dashed vertical lines show what amount of overlap corresponds to human performance in Potter and Thorpe experiments. }
\label{fig4}
\end{figure}

To better understand the appropriateness of these choices, we conducted a sensitivity analysis for each of the algorithms and this is shown in Fig~\ref{fig4}.  The plots show how the percentage of points of interest (P) within the ground truth masks and parafovea gradually decreases depending on how much of the GTM (by area) is inside  the parafovea. On each plot the point (0,0) corresponds to measure B (i.e. P is within the GTM and parafovea regardless of the amount of GTM within the parafovea). Dashed vertical lines show what amount of overlap corresponds to human performance in Potter and Thorpe experiments. The overlap of 27\% corresponds to measure C and 41\% to measure D shown in Fig~\ref{fig5} (C and D).  This analysis makes clear that the threshold choices are indeed sensible and fair, and that they are not sensitive to small changes. Finally, as will be shown in the Results section 'c', these threshold choices are justifiable through human experimentation.

\subsection*{Algorithm biological plausibility}
Although many computer vision algorithms take significant inspiration from the human visual system, few embody that inspiration in a biologically realistic manner. It is difficult to evaluate such algorithms as to their biological plausibility but there is at least one tool that can be used. Feldman \& Ballard \cite{feldman1982connectionist} explicitly linked computational complexity to neural processes saying ``Contemporary computer science has sharpened our notions of what is `computable' to include bounds on time, storage, and other resources. It does not seem unreasonable to require that computational models in cognitive science be at least plausible in their postulated resource requirements.'' They go on to examine the resources of time and numbers of processors, and more, leading to a key conclusion that complex behaviors can be carried out in fewer than one hundred (neural processing) time steps. These time steps were considered to be roughly the time it might take a single neuron to perform its basic computation (coarsely stated as a weighted sum of its inputs followed by a non-linear transformation) and then transmit its results to the next level of computation, perhaps about 10ms. Thorpe \& Imbert \cite{thorpe1989biological} also place similar constraints on processing time and numbers of layers suggesting that at least 10 layers of about 10msec per layer are needed.  Combining this with Thorpe et al.'s observation that 150ms suffices for yes-no category decision, this constrains biologically plausible algorithms to those requiring no more than 15 or so layers of such computations. Since the algorithms we are testing do not deal with the full problem of categorization but only reflect the saliency computation stage, one might expect a much smaller time constraint, i.e., significantly fewer than 15 layers of computation.

\subsection*{Human categorization performance to parafoveal stimuli}

If early selection, salience-based or otherwise, were important for human rapid categorization performance, then testing humans with images where only the region within an observer's parafovea (within $2.5\,^{\circ}$ radius of point of fixation) would be revealing. Good performance would show early guidance is unnecessary. In other words, the current pre-determined fixation would suffice for good performance implying a shift in fixation would not lead to meaningful improvement. We took the original image set of Thorpe et al. and cropped each image to its parafovea content and tested subjects for categorization performance.

\section*{Results}

\subsection*{Algorithm performance}

\begin{figure}[!h]
\centering
\includegraphics[width=0.7\linewidth]{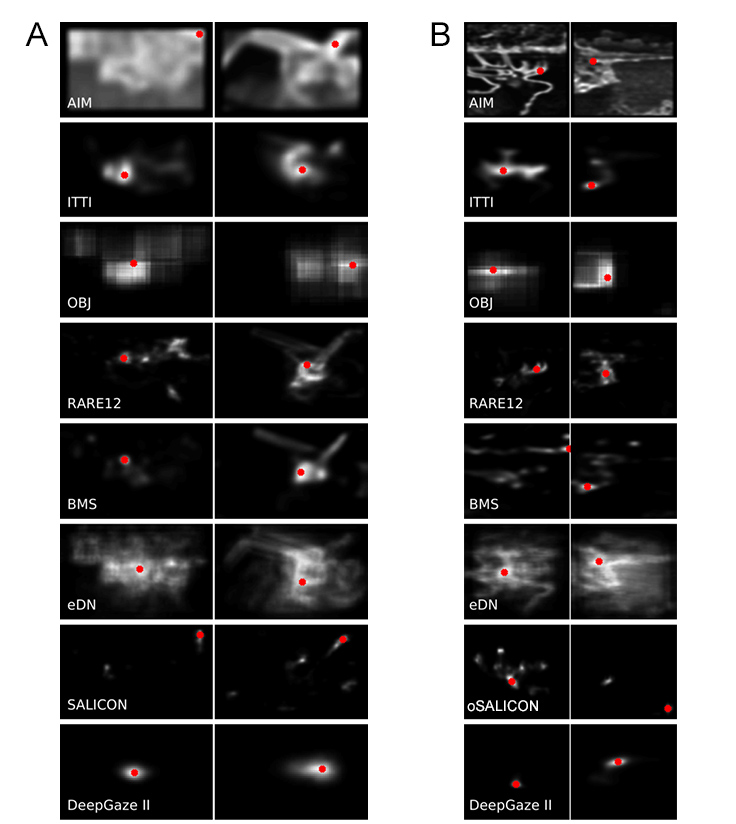}
\caption{{\bf Examples of saliency maps.} Saliency maps generated by the 8 saliency algorithms for sample images from the Potter (A) and Thorpe (B) datasets shown in Fig~\ref{fig1} with target present (left column) and no target present (right column). The red dot in each saliency map marks the
global maximum found in the saliency map (the most likely first attended location predicted by the algorithm).
}
\label{fig5}
\end{figure}

Fig~\ref{fig5} shows one example saliency heat map from each of the two datasets for each algorithm, one for a target-present and one for target-absent (the images used are those shown in Fig~\ref{fig1}). 

\begin{figure}[!h]
\includegraphics[width=\linewidth]{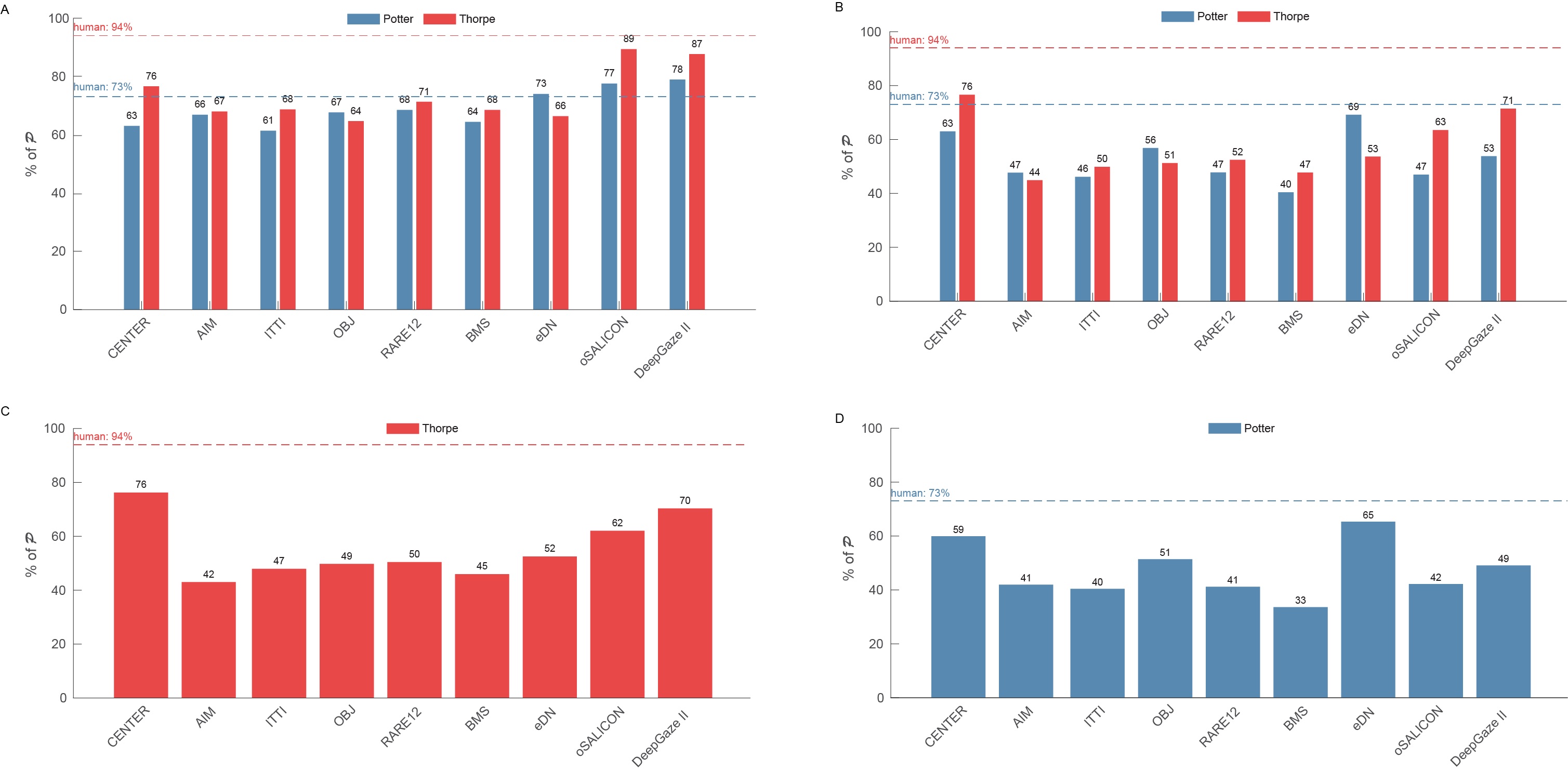}
\caption{{\bf Plots of results using 4 performance measures.} A: The percent of all first $\mathcal{P}$ that fall anywhere within the GTM for each tested algorithm and dataset. B: The percent of all first $\mathcal{P}$  that fall that fall within the GTM AND within the parafovea for each tested algorithm and dataset. C: The percent of all first $\mathcal{P}$ that fall within the GTM AND within the parafovea AND at least 27\% of the GTM (by area) lies within the parafovea for each algorithm but for only the Thorpe images. D: The percent of all first $\mathcal{P}$ that fall within the GTM AND within the parafovea AND at least 41\% of the GTM (by area) lies within the parafovea for each algorithm but only for the Potter images.
}
\label{fig6}
\end{figure}

The overall categorization performance data is seen in Fig~\ref{fig6}, where parts A, B, C and D correspond to the four performance measures described earlier, respectively.  

This test reveals several results:
\begin{enumerate}[label=\alph*)]
	\item{Using the most generous measure, A, several algorithms approach  human level performance.}
	\item{Using the more appropriate measure, B, only eDN approaches human performance on the Potter set and no algorithm comes close on the Thorpe dataset (even if the human performance level is reduced by 10\% to compensate for prior instruction).}
	\item{Using the measure C tailored for the Thorpe dataset, eDN leads the pack but again, somewhat below human performance.}
	\item{Using the measure D tailored for the Potter dataset, DeepGaze II is closest, but quite below human performance.}
	\item{Interestingly, the CENTER algorithm works almost as well as the best algorithms and sometimes outperforms all methods.}
\end{enumerate}

\begin{figure}[!h]
\includegraphics[width=\linewidth]{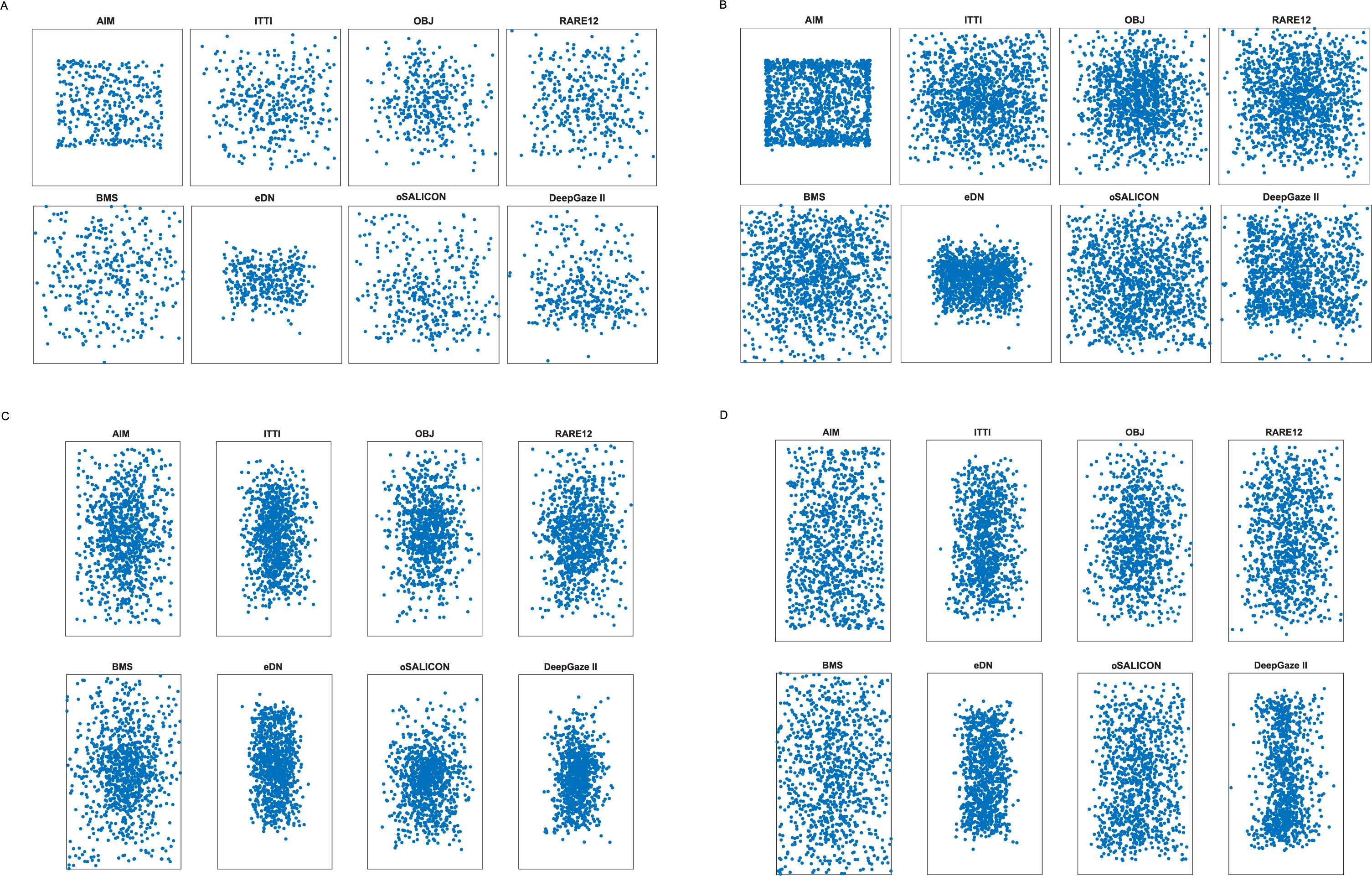}
\caption{{\bf Scatterplots of the first attended locations $\mathcal{P}$ predicted by the saliency algorithms.} A: $\mathcal{P}$ for target-present images in the Potter set. B: $\mathcal{P}$ for images with no target present from Potter set. C: $\mathcal{P}$ for target-present images in the Thorpe set. D: $\mathcal{P}$ for images with no target present from Thorpe set. 
}
\label{fig7}
\end{figure}

While performing these metric tests, we also plotted the locations of all the $\mathcal{P}$'s and these scatterplots are shown in Fig~\ref{fig7}. It is easy to notice that there was a center bias in some cases as well as issues with boundary effects. Fig~\ref{fig6} shows scatterplots of algorithms' first $\mathcal{P}$ location for the Potter dataset in parts A (target present) and B (target absent) and the same for the Thorpe dataset in parts C and D. 

The biases are striking. The AIM algorithm clearly has a problem with image boundaries. It might be ameliorated through the use of image padding as other algorithms employ, although this is not a biologically realistic solution. ITTI, eDN, oSALICON and DeepGaze II also seem to have a preference for more central $\mathcal{P}$ results. It was shown earlier that the stimulus image sets do contain center bias for their targets; however, these algorithms demonstrate a center bias for the no-target cases as well. Even though we turned off explicit center bias computations for ITTI, eDN and DeepGaze II, it seems that these algorithms have additional implicit center biases. Their good performance for the target-present cases is perhaps suspect as a result.

\subsection*{Algorithm biological plausibility}
Although many computer vision algorithms take significant inspiration from the human visual system, few embody that inspiration in a biologically realistic manner and most include extensions and enhancements in an attempt to outperform humans. It is difficult to evaluate such algorithms as to their biological plausibility but there is at least one tool that can be used. Feldman \& Ballard \cite{feldman1982connectionist} explicitly linked computational complexity to neural processes, and Thorpe \& Imbert \cite{thorpe1989biological} further add to this as described earlier.  We thus constrain biologically plausible algorithms to those requiring no more than 15 or so layers of neural computations. Since the algorithms we are testing do not deal with the full problem of categorization but only reflect the saliency computation stage, one might expect a much smaller time constraint, i.e., significantly fewer than 15 layers of computation.

Table~\ref{table1} gives a coarse evaluation of the number of levels of computation, using the approximate criteria just described, for each of our tested algorithms. The AIM, BMS, ITTI, and eDN algorithms seem well within the timing constraints stated, with DeepGaze II and oSALICON outside the constraint of significantly fewer than 15 layers. Of those that do fall within the time step constraint, only eDN shows good performance, primarily on the Potter set. The depth and style of computation of the DeepGaze II and oSALICON algorithms mimics a full feedforward pass through the visual hierarchy. This perhaps argues for an incremental selection process which would be a valid possibility in human processing as well, although none of the models cited in this paper consider it (note Treisman's attenuated selection idea \cite{treisman1964effect}).

\begin{table}[!ht]
\begin{adjustwidth}{-2.25in}{0in} 
\centering
\caption{
{\bf Number of neural processing layers in saliency algorithms}. For each of the tested algorithms, an estimate of the number of neural-equivalent processing layers is presented. The AIM, BMS, ITTI, eDN algorithms seem within the timing constraints stated, with oSALICON and DeepGaze II outside.}
\begin{tabular}{|l|l|l|}
\hline
\multicolumn{1}{|c|}{\textbf{Algorithm}} & \textbf{Processing steps}                                                                                                                                                                                                                                                                                                                                            & \textbf{\begin{tabular}[c]{@{}l@{}}Depth\end{tabular}} \\ \hline
AIM  & Feature filter $\rightarrow$ Density estimation $\rightarrow$ Self-information                                                                                                                                                                                                                                                                                       & 3                                                                \\ \hline
BMS  & \begin{tabular}[c]{@{}l@{}}Feature split $\rightarrow$ Threshold feature channels $\rightarrow$ Connected components + Normalization $\rightarrow$ \\Average + Dilation\end{tabular}                                                                                                                                                                             & 4                                                                \\ \hline
ITTI & DoG and Gabor Filters $\rightarrow$ Average features + Normalization $\rightarrow$ Average channels                                                                                                                                                                                                                                                                  & 3                                                                \\ \hline
eDN & Ensemble of CNNs (max 3 layers deep) $\rightarrow$ SVM combination of CNN output                                                                                                                                                                                                                                                                                     & 4                                                                \\ \hline
OBJ & \begin{tabular}[c]{@{}l@{}}4 parallel streams: {[}SR saliency - depth1; Patch colour contrast - depth 1;\\Edge detection $\rightarrow$ Edge counting - depth 2; Superpixel straddling - depth 5{]} $\rightarrow$ \\Bayesian integration{]}\end{tabular}                                                                                                      & 6*                                                                \\ \hline
RARE2012 & \begin{tabular}[c]{@{}l@{}}PCA colour decomposition $\rightarrow$ log-Gabor filtering $\rightarrow$ Averaging and normalization of\\ log-Gabor scales $\rightarrow$ Gaussian pyramid + Density estimation $\rightarrow$ Self-information of channels $\rightarrow$\\ Weighted average within channels $\rightarrow$ Weighted average between channels\end{tabular} & 7                                                                \\ \hline
oSALICON & Two-stream VGG-16 fine-tuned to saliency detection                                                                                                                                                                                                                                                                                                                   & 16                                                               \\ \hline
DeepGaze II  & Extract features from VGG-19  $\rightarrow$ readout network                                                                                                                                                                                                                                                                                                          & 19                                                               \\ \hline
\end{tabular}
\begin{flushleft}* The equivalent convolutional depth of the superpixel step is taken to be $log(n)$ convolutions (where n is the number of pixels in the image), which works out to be approximately 5 layers for the images dealt with here.
\end{flushleft}
\label{table1}
\end{adjustwidth}
\end{table}

\subsection*{Human categorization performance to parafoveal stimuli}
The original stimulus set was revisited for human categorization performance to test the hypothesis that humans do not require the full image to achieve their high level of performance, and perhaps only the portion of the stimulus that is seen in an observer’s parafovea was required. If the test images for rapid human categorization are center-biased as was demonstrated earlier for both the Thorpe and Potter datasets, this means there is little need for humans to require a shift in ROI if the subject’s parafovea is at the image center since the target is usually right there too. This is necessarily true since subjects are instructed to maintain a center gaze. One can thus ask whether the parafovea is a sufficient ROI so that there would be no need to adjust its position in order to obtain good performance. We asked several questions:

\begin{enumerate}
	\item{What is the relationship between accuracy of categorization and the portion of the target that falls within the parafovea?}
	\item{If the test images are cropped to be only the portion within the parafovea (that is, a circular region with $2.5\,^{\circ}$ radius), what is human categorization performance?}
	\item{What is the relationship between accuracy and portion of target within the parafovea for the 3 top performing algorithms (oSALICON, eDN and DeepGaze II)?}
\end{enumerate}

\begin{figure}[!h]
\includegraphics[width=\linewidth]{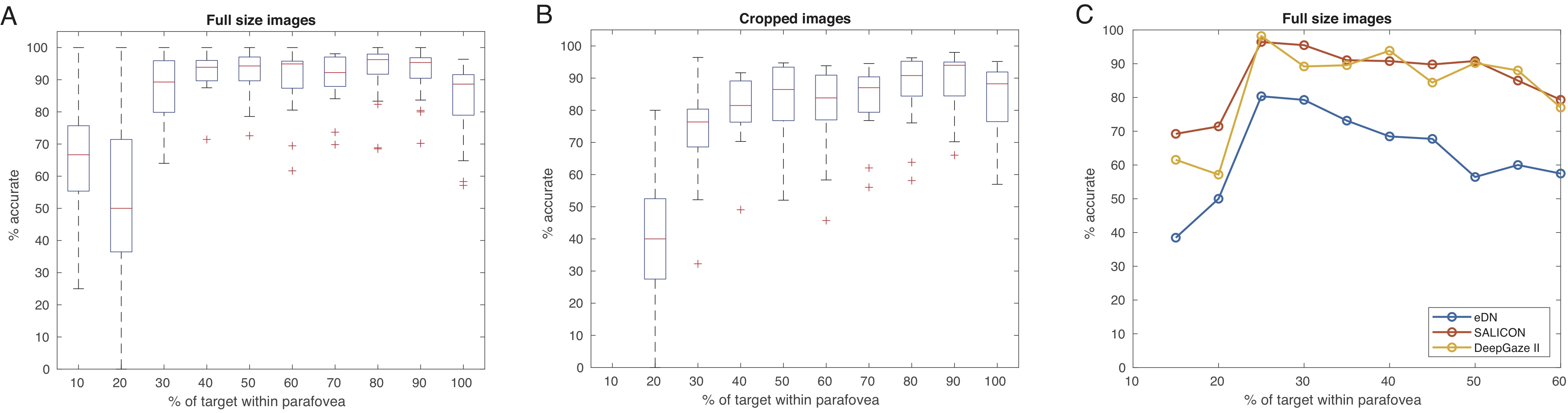}
\caption{{\bf Results of human and algorithmic categorization performance.} A,B: Box plots of human categorization performance plotted as a function of the percentage of the target within the parafovea. In the ``Full size'' condition (A) we used the full, original images in Thorpe's dataset. In ``Cropped'' condition (B) images from Thorpe's dataset were cropped so that the image portion outside the parafovea is set to a light grey. Recall that Thorpe reported 94\% accuracy across the full image set. C: Performance of the top 3 saliency algorithms (oSALICON, eDN and DeepGaze II). The plot shows the percentage of correct responses (vertical axis) vs the percentage of the target within the parafovea.
}
\label{fig8}
\end{figure}

The results are shown in Fig~\ref{fig8} respectively (the experimental procedure is detailed in \nameref{S3_Text}). The central red mark within each box indicates the median, and the bottom and top edges of the box indicate the 25th and 75th percentiles, respectively. The whiskers extend to the most extreme data points not considered outliers, and the outliers are plotted individually using the `+' symbol. Fig~\ref{fig8}C shows performance of the top 3 saliency algorithms. We used measure A to compute average accuracy on images within bins representing \% of target covered by parafovea.

Our experiment roughly duplicates Thorpe’s results for the full, original images using his experimental parameters and protocol, as Fig~\ref{fig8}A shows. High performance corresponded to at least 30\% of the target being present in the parafoveal region. Even when the images are cropped to blank out extra-parafoveal portions, performance remained quite high as long as 30-40\% of the target was inside the parafovea as seen in Fig~\ref{fig8}B. Algorithm performance showed the same characteristic with the full images in Fig~\ref{fig8}C; high accuracy was obtained whenever over 30\% of the target was present in the parafovea. In other words, to a good approximation, both algorithm and human performance can be predicted by \% target in the parafovea; setting a ROI within the image via any method would play little or no role. In addition, these results provide justification for our definition of performance measures C and D given earlier; for the Thorpe images, we had defined a positive hit if $\mathcal{P}$ falls within both the GTM AND the parafovea AND at least 27\% of the GTM (by area) lies within the parafovea. From the results in Fig~\ref{fig8}A, a figure perhaps closer to 30-35\% might have been better. The definition we used was thus generous in favor of the algorithms.

\section*{Discussion}
The analysis just presented is relevant only to the role saliency computation might play for human vision. It does not directly address the role of saliency in computer vision; however, it does highlight the fact that the use of saliency computation in computer vision might have no biological motivation or justification. In machine vision, it might well play an important role in limiting the extent of the image that needs to be processed by selecting more relevant image portions (the arguments from computational complexity described in \cite{tsotsos1988complexity}, \cite{tsotsos1989complexity}, \cite{culhane1992attentional}, \cite{tsotsos1995modeling} conclude exactly this point). Within the set of saliency algorithms examined, we also uncovered some interesting, and not previously revealed, biases that certainly affect their overall performance. The ITTI, eDN, oSALICON and DeepGaze II algorithms seem to have a preference for more central $\mathcal{P}$ results (even after the built-in center bias was turned off for ITTI, eDN and DeepGaze II, while oSALICON does not include an explicit center prior). This places some doubt on their strong performance for the target-present images. The rest of this section will focus on several specific points of the analysis. The overall conclusion is that neither classic nor modern saliency algorithms, in their supporting role of ROI prediction, would lead to the same high level of categorization performance in humans. Further, a closer analysis of human performance shows that there is little point to an accurate feedforward point or region of interest prediction in the original experiments.

Firstly, it is important to explain why it is justified to test existing saliency algorithms in this manner when none of them can accept nor use task specifications or prior instruction. The only Potter results we use as comparison are those without subjects receiving prior guidance, while those we use of Thorpe included uniform guidance for expected category. It has already been stated that in the Potter et al. experiment \cite{potter2014detecting} without prior expectation, accuracy was roughly 10\% lower than with prior guidance. van der Heijden et al. \cite{van1985enhancing} also reached a similar and thus consistent conclusion, finding a 12\% relative change in performance between spatial cue present and absent conditions. It thus seems reasonable to assume a similar decrease in the Thorpe experiment. That is, accuracy would likely be only about 10\% lower if subjects had no instruction as to category. If we reduce the Thorpe image set human performance by 10\%, then in Fig~\ref{fig6}A oSALICON and DeepGaze II exceed human performance, but the conclusions from Fig~\ref{fig6}B and Fig~\ref{fig6}C remain the same. Additionally, both algorithms do not seem to fit within the computational layer constraints. Finally, in Fig~\ref{fig8}C, algorithm performance is shown comparable to human performance for sufficient target presence within the parafovea, i.e., the saliency prediction is not needed.

There is a problem with the set of metrics we used that should be acknowledged. Following measures C or D for the elk image of Fig~\ref{fig3}, none of the algorithms tested would give a correct hit if the predicted P falls anywhere within the red region in the figure. This is not because of any fault of the algorithm but rather the definition of the metric itself which is tied to the position and scale of the target with respect to the observer’s parafovea. This will lead to some amount of under-estimation of the accuracy of the algorithms, but the goal was to estimate how much of the target should be in the parafovea in order for a human to recognize it at the observed levels. It has already been acknowledged that the measure was not completely accurate. On the other hand, one can imagine an image where the target is small and completely within the parafovea, where a saliency algorithm hits it directly, but is not recognized by the observer because it is too small or may lie within distracting background elements. This would over-estimate accuracy with respect to humans. These two tendencies may balance each other to some degree. Nevertheless, our human experiments, shown in Fig~\ref{fig8}, inform us that the assumptions made in defining the measures C and D are reasonable.

We return to the issue of accuracy measures used now looking at them from a signal detection perspective. The accuracy measures reported by Thorpe and colleagues represent the averaged sum of True Positives plus True Negatives (TP+TN). We were thus constrained in our comparison wishing to align our conclusions to the human performance they reported. However, a later paper from Thorpe's group does provide an opportunity for a better signal detection analysis, with the difference that the experimental subjects are rhesus monkeys rather than humans.  Our own human experimental data is also amenable to this more complete analysis.

Fabre-Thorpe et al. \cite{fabre1998rapid} considered rapid categorization tasks of natural images by rhesus monkeys. They note that the task presented to their subjects used the same stimulus types as the Thorpe et al. work \cite{thorpe1996speed}, with similar methods (including additional direct human tests), and they observed similar results, leading them to conclude that humans and monkeys likely use very similar processes for these tasks. The fact that the two sets of experiments were preformed in the same lab adds credibility to their assertion. It is therefore reasonable to compare our results to this paper. In contrast to the Thorpe et al. paper \cite{thorpe1996speed} where only accuracy is reported, this later paper provides a fuller report of performance. The first group of  rows of Table~\ref{table2} gives the results from Fabre-Thorpe et al. \cite{fabre1998rapid} while the middle group of rows give our experiments (described earlier).

\begin{table}[]
\caption{\textbf{Comparison of human and monkey categorization performance with the Ideal Decision Stage results.} The table provides a comparison of performance for the Fabre-Thorpe et al. experiment \cite{fabre1998rapid}, our own human experiments, and the Ideal Decision Stage we have assumed. Accuracy is computed as (TP+TN)/P+N.  The TP entry for the Ideal Decision Stage reflects the best performing saliency algorithm from Fig~\ref{fig6}C. Recall that the Thorpe et al. paper \cite{thorpe1996speed} reported average human accuracy of 94\% correct. }
\begin{tabular}{ll|c|c|c|c|c|}
\cline{3-7}
\multicolumn{1}{c}{}                                                                                   & \multicolumn{1}{c|}{} & Accuracy & TP   & FN   & TN   & FP   \\ \hline
\multicolumn{1}{|l|}{\begin{tabular}[c]{@{}l@{}}Fabre-Thorpe et al.\\ monkey experiments\end{tabular}} & New Images            & 0.84     & 0.99 & 0.01 & 0.69 & 0.31 \\ \cline{2-7} 
\multicolumn{1}{|l|}{}                                                                                 & Familiar Images       & 0.89     & 0.96 & 0.04 & 0.83 & 0.31 \\ \hline
\multicolumn{1}{|l|}{\begin{tabular}[c]{@{}l@{}}Our human \\ experiments\end{tabular}}                 & Full Images           & 0.93     & 0.94 & 0.06 & 0.93 & 0.07 \\ \cline{2-7} 
\multicolumn{1}{|l|}{}                                                                                 & Cropped Images        & 0.85     & 0.88 & 0.12 & 0.85 & 0.15 \\ \hline
\multicolumn{1}{|l|}{\begin{tabular}[c]{@{}l@{}}Ideal Decision \\ Stage\end{tabular}}                  & Full Images           & 0.85     & 0.70 & 0.30 & 1.00 & 0.00 \\ \hline
\end{tabular}%
\label{table2}
\end{table}

For our saliency computations each algorithm produced a point of interest regardless of image content so it is not possible to make a direct comparison. Let us assume an Ideal Decision Stage (IDS). This stage receives the fixation point from a saliency algorithm, knows what the target is, and then always outputs the correct conclusion for that point. If the fixation prediction lies within the target object (as defined by our measure C), then the output of IDS is always 'yes'. For the target-absent cases, the output will always be 'no'; it does not matter where the fixation point is, it never points to a target. In other words, for target-present trials, the output will be 'yes',  a True Positive if the P is close enough to the target centroid (measure C). If P is not close enough, then the IDS will yield a 'no', a False Negative. For target-absent trials, the output of the IDS will be 'no', thus a True Negative, so TN=1.00. There is no possibility of a False Positive since this is an ideal decision, so FP=0 always. These are entered in the final row of Table~\ref{table2} which provides the performance of the IDS. It should be clear that even with the Ideal Decision Stage assumption, saliency algorithms do not approach human nor monkey performance.

From the saliency algorithm point of view, the algorithm believes it has a potential target for all trials and does not discriminate between target present or absent scenarios. The ideal decision stage, of course, does not know which scenario is being presented. However, since it processes only the point/region of interest, it necessarily would make errors for target-present cases where the saliency algorithms produce poor predictions but will always be correct for target-absent cases, because there could be no match to a target even though it would not have verified the absence of the target by examining the whole image.  As our tests show, existing saliency algorithms do not reach the level of human accuracy and as a result, the detector's upper bound for True Positives would be the same as that of the saliency algorithms (whose best correct performance for target-present trials is 70\% according to Fig~\ref{fig6}C; thus TP=0.70 and FN=0.30 in the table).  The saliency algorithms always produce a fixation prediction; however, in this categorization context this means that they always produce a misleading prediction for the ideal decision stage for target-absent trials. Since the whole process  occurs within 150ms for both 'yes' or 'no' output, and stimulus exposure is so short, there is no time for additional processing.  That is, there is no time to test that prediction, and once it is confirmed that it does not include a target, to try again until the full image is checked. Thus the Ideal Decision Stage will always be correct: the stimulus has no target. Saliency suggests a predicted fixation, it is checked and rejected. But it would be for the wrong reason. Since only one ROI is checked the overall system cannot be certain. Our assumption for the Ideal Decision Stage applies only to that stage not to the whole system. Note that  human performance on target-absent cases is not perfectly accurate (see Table~\ref{table2}) so it seems that humans do not always correctly check the entire image either.

It is reasonable to consider what would be appropriate for the output of a saliency algorithm for the target-absent stimuli. Firstly, it seems important for the algorithm to include knowledge of what the target is in order for it to able to distinguish targets from non-targets. The classic saliency definition has no such component; it is purely a feedforward computation depending on local image contrast alone. Even if the definition changed to include such a top-down component, an output based on a maximal local contrast computation would not suffice. For a target-absent stimulus, some separate, more global, computation seems required, perhaps of the type  argued by Herzog \& Clarke \cite{herzog2014vision}. Could it suffice to use a variance detector that gives a global measure of low variability across an image if there is no target? This is unlikely without knowledge of the target affecting the determination. In any case, this alone could not tell the difference between an image with only regions of low interest (i.e., low local contrast) and an image with many salient regions. It would seem that some absolute measure of saliency is needed rather than a relative one (something which is impossible to do when saliency maps normalize their output as standard practice). These characteristics no longer come close to the definition nor practice of saliency computations as seen between 1985 and the present. They may however, point to directions for future computational as well as human experimental work. This potential notwithstanding, the work reported here means that it is highly unlikely that a strictly feedforward and spatially local process - as the early selection concept dictates - can suffice to drive human rapid visual categorization.

If early selection guides the process in humans, and since there is no time to check more than one ROI, then one might think that there might be some other way for checking the whole image. Perhaps humans use an entirely separate parallel stream that not only takes at most 150 ms, but considers the full image (see the 2nd paragraph of the discussion in the 1998 Fabre-Thorpe et al.  paper \cite{fabre1998rapid} where they argue against a sequential component to this task). The second parallel process needs to be "on" always - there is no controller to decide if the saliency stage output is valid or not before deploying it - there is no time for this. So if it is always on, then a decision stage is needed to decide which output is the one to report  - the one guided by saliency or the one not so guided. Since the one not guided makes a global determination, and the one guided by saliency only a local one, the global one should always be preferred,  if the system is a rational agent. But all this  brings us back to the original hypothesis - a single feedforward pass that takes 150ms and has the full image as input where saliency plays no role. 

It is certainly true that a better saliency algorithm that fits within the strategy first outlined by Koch \& Ullman \cite{koch1987shifts} may yet be discovered. A still different possibility could be that even within the parafovea, some kind of early selection is taking place perhaps performing a tentative figure-ground segmentation and that selected figure is then passed on for further processing. But this seems to be simply changing the scale of the image; early selection within the fovea still has the same problem - how to be certain that a target is not elsewhere within the parafovea, and thus, this possibility does not suffice to solve the problem. With either possibility, the fact that the target-absent images would remain incompletely processed remains.

\section*{Conclusions}
The widely held position that visual saliency computation occurs early during the feedforward visual categorization process in human vision was tested and we found no support for it. It is emphasized that the conclusion applies only to human vision and not any saliency role useful for machine vision systems. In fact, our experiments have shown that many machine algorithms, freed from anatomical or resource constraints that bind the human visual system, perform very well. 

This is not to say that saliency computation has no role in any other aspect of human vision. In Tsotsos et al. \cite{tsotsos2016focus} we describe a novel eye fixation prediction algorithm that employs several forms of saliency computation but not as selection for categorization tasks. It is a hybrid model that combines the positive elements of early selection, late selection, and more. We provide arguments that a cluster of conspicuity representations drives eye fixation selection, modulated by task goals and fixation history. Quantitative evaluation of this proposal shows performance that falls within the limits of human performance evaluation, and is far superior to any of the saliency methods tested \cite{wloka2018active}.  Thus, visual saliency has at least the important role of participating in eye fixation computations.

Our experiments have shown that no tested algorithm can provide a sufficiently accurate first region-of-interest prediction to drive categorization results at human behaviour levels. In fact, little is gained by all the effort in comparison to the CENTER control model we tested. The many models and theories of human visual information processing, although inspiring and useful for many years of research, have served their role as important stepping stones on the path to understanding vision, but now may need to be reconsidered. Those saliency algorithms which do approach human performance seem too computationally expensive to also be biologically plausible as early selection mechanisms. It should be noted that the computational expense is not so large as to make them completely implausible; they could point to a continuous or incremental selection mechanism (as opposed to early or late) and this might be an interesting direction for future exploration. However, there is no provision in any algorithm for the target-absent stimuli. They cannot provide a ‘no target’ result in the same processing time; the salience-based processing strategy forces a serial search without the global computation a correct target-absent conclusion requires. We also tested human visual categorization and found that human performance seems to not need early salience. It appears sufficient for good categorization that some reasonable amount of the target appears in a subject’s parafovea. Human performance was strongly predicted simply by the spatial relationship between target and the observer’s parafovea, leading to the conclusion that a region-of-interest derived by any means adds little to human performance for conditions where gaze is fixed on the image center.
\newpage
\section*{Supporting information}


\begin{figure}[!h]
\includegraphics[width=\linewidth]{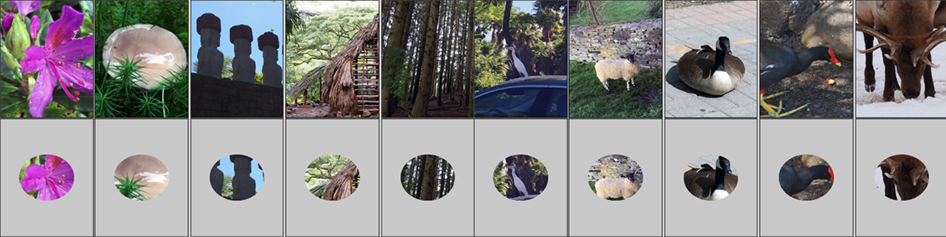}
\caption{{\bf Sample images from the Thorpe dataset with and without targets (animals).} The top row shows full images used in the first experimental condition and the bottom row shows the same images cropped to the size of the parafovea, used in the second experimental condition.
}
\label{Fig9}
\end{figure}

{

\paragraph*{S1 File.}
\label{S1_File}
{\bf Experiment data.} Mean accuracy values for 17 human subjects and 3 saliency algorithms. Values are aggregated over different percentages of object within the parafovea region (between 10\% and 100\% of target by area). Data can be downloaded at: \url{https://doi.org/10.1371/journal.pone.0224306.s002}.

\paragraph*{S1 Text.}

\label{S1_Text}
{\bf Saliency models.} The code for all saliency algorithms that we used is publicly available and was not modified for the experiments. Default parameters were used in each case. Below is a list of links. 
\begin{description}
\item[AIM] \url{https://github.com/TsotsosLab/AIM}.
\item[ITTI]	implementation provided with the GBVS saliency package 	(\url{http://www.vision.caltech.edu/~harel/share/gbvs.php}).
\item[OBJ] v2.2 of the algorithm (\url{http://groups.inf.ed.ac.uk/calvin/objectness/}).
\item[RARE2012] \url{http://www.tcts.fpms.ac.be/attention/?categorie17/rare2012}.
\item[BMS] we use the most recent version for eye-fixation prediction  from \cite{zhang2016exploiting} (\url{http://cs-people.bu.edu/jmzhang/BMS/BMS.html}).
\item[eDN] \url{https://github.com/coxlab/edn-cvpr2014}
\item[oSALICON] since the original algorithm is not published, we use the open source version of it that has comparable results as described in \cite{thomas2016opensalicon}	(\url{https://github.com/CLT29/OpenSALICON}).
\item[DeepGaze II] \url{https://deepgaze.bethgelab.org/} (the code was not available at the time of the writing. We used the saliency maps provided by the authors to our data).
\end{description}

In all experiments we used default parameters. 

OBJ outputs multiple object proposals as bounding boxes. These bounding boxes then can be combined to form a heatmap that approximately corresponds to a saliency map. The latter option was used in the experiments.

In order to evaluate inherent center bias of the algorithms we turned off the explicit center prior used in the following saliency models: ITTI, eDN and DeepGaze II. Other algorithms in our selection do not use the explicit center prior.

Even though we switched off the explicit center prior where applicable, there is still a possibility that some models, in particular the deep learning ones, may learn center bias from the training data. For example, the OSIE dataset \cite{xu2014predicting} (used to train the oSALICON model), the oSALICON dataset \cite{jiang2015salicon} (used for the DeepGaze II model) and the MIT1003 dataset \cite{judd2012benchmark} (used for eDN) all have significant center bias.

\paragraph*{S2 Text.}
\label{S2_Text}
{\bf Human retina characteristics.} Our focus on the parafoveal region of a test image can be justified by considering the following. Sources of relevant information on photoreceptor distribution and other retinal characteristics in humans include \cite{osterberg1935topography}, \cite{curcio1990human}, and \cite{curcio1990topography}. Without recounting all the details, it is well-known that density of retinal cones is at its maximum at the very center of the fovea and falls rapidly towards the periphery. At its center lies the foveola, 350$\mu$m wide ($0.5\,^{\circ}$) that is totally rod-free and capillary free, thus seeming the optimal target for new visual information. The parafovea is the region immediately outside the fovea with a diameter of 2.5mm ($5\,^{\circ}$). It is important to recall that acuity decreases with retinal eccentricity. Anstis \cite{anstis1974chart} showed that to maintain visual acuity an object must increase by 2.76 arcmin in size for each degree of retinal eccentricity up to about $30\,^{\circ}$, and then somewhat more steeply up to $60\,^{\circ}$. It seems clear that if the target object that falls within the central $5\,^{\circ}$ of the retina, the likelihood of its correct categorization is much higher than if otherwise.

\paragraph*{S3 Text.}
\label{S3_Text}

{\bf Experimental Details for the Parafoveal Stimuli Test.}

\noindent
\textit{Procedure}. There are two conditions in the study. In the first condition, we replicate the experiment conducted by Thorpe et al. \cite{thorpe1996speed}. In the experiment subjects view an image for the duration of 20ms and perform a go/no-go categorization where they have to decide whether an animal is present in the scene or not. Since no specific task definition was provided in the original description of the experiment, we have asked our participants to look for a live animals excluding humans. In addition, the subjects were instructed to ignore artistic renditions of animals, such as those corresponding to drawings, statues, etc.

In the second condition, the images were cropped so that only the area within the circle with radius of $2.5\,^{\circ}$ (corresponding to the size of the parafovea) remained visible. Subjects were instructed to maintain fixation on the cross presented and to not move their eyes. Each trial begins with a fixation cross for 500ms, followed by an image for 20ms and a fixation cross, on a blank screen, for 500ms. The images are shown consecutively with a random interval of 1 to 2 seconds during which subjects have to press the space bar on the keyboard if they see an animal in the image.

\textit{Participants.}
A total of 17 subjects (6 women, 11 men), between the age of 25 and 34 years old, participated in the study. All participants were volunteers and were not compensated for their participation. Additionally, the participants were asked to sign a consent form approved by the York University Office of Research Ethics (Certificate number 2016-014 ``Selective tuning approach to visual system attention executive''). Each subject completed 10 blocks of 100 images for each condition. 

\textit{Materials.}
The stimuli were 2000 color photographs as used in the original experiment in Thorpe et al. The subjects were not familiar with the images and viewed each image only once. For each participant, the data was randomly split into two equally-sized sets for each experimental condition containing approximately the same number of images with and without animals. The images were resized to 256 by 384 pixels and were presented in the center of the monitor on a light gray background.

\textit{Apparatus.}
The experiments were programmed in Matlab R2016b using the Psychophysics Toolbox (Brainard 1997) version 3. The monitor, a ViewSonic Graphics Series GS815 19 in CRT, was set to 1024 x 768 resolution with 75 Hz refresh rate. All subjects were placed in a dark room and were seated 60 cm away from the monitor with their head movements restricted by a chin rest.

\textit{Results.}
The results of the first condition (full images) are similar to the ones reported by Thorpe et al. The average proportion of correct responses is 93\% compared to the 94\% in the original experiment. One of the subjects achieved the rate of 97\% correct responses (98\% in Thorpe et al.). In the second condition (cropped images) 85\% of responses are correct with a maximum of 91\% achieved by one of subjects. In the second condition we excluded trials where the target was located outside the parafovea region (overall $<$1\% of the trials were removed as a result). Examples of full-size and cropped images used in the experiment are shown in Figure \ref{Fig9}.

We analyzed individual participants’ percentages of correct responses as a function of the percentage of overlap between the parafovea mask (r=$2.5\,^{\circ}$) and a binary mask corresponding to the animal in the image. Only responses on target-present trials were considered for this analysis (mean responses of human subjects are provided as S1 File). 

Examples for the full-size and cropped images are shown in top and bottom row of Figure \ref{Fig9} respectively. The results of the first experimental condition are shown in Fig 8A. Note that despite significant differences in individual performance, overall, most subjects have lower response accuracy when the target covers between 10 and 30\% of the parafovea. The accuracy levels out when the target occupies $>$40\% of the parafovea. 

There is much more variability in the subject responses in the second experimental condition (cropped images), shown in Fig 8B,  because most of the context is not available (see Figure \ref{Fig9} bottom row). However, the same trend as in the first condition is still very noticeable. For instance, when the target occupies $>$20\% of the parafovea, the average response accuracy is above 60\%. Particularly, larger targets, covering $>$70\% of the parafovea, were challenging because most of the animal was likely to be cropped out. Furthermore, in many cases, easily identifiable parts of the animals, such as head, wings, antlers, etc., are not necessarily present in the central region and for the images where the targets covered 100\% of the parafovea, human/algorithm performance was a little worse. 

We conducted a repeated-measures analysis of variance (ANOVA) on the human experimental data. The effect of target overlap with the parafovea on the accuracy of responses was significant in both conditions: F(9, 144)=76.784 $<$ 0.001 and F(9, 126)=100.305 $<$ 0.001 respectively.

Fig 8C shows the performance of the top 3 saliency algorithms (oSALICON, eDN and DeepGaze II). Note that this plot is only provided for qualitative comparison since it was not possible to subject the algorithms to the same experimental conditions as human participants. For instance, response times are not comparable due the fact that humans are required to make motor responses. Furthermore, we generously assume that the target is recognized by the algorithm if the maximum of the saliency map falls within the ground truth mask (Measure A). Therefore, results shown in Fig 8C should be interpreted as the upper bound on the accuracy of the saliency algorithms. 

Several observations can be made based on the box plots shown in Fig 8. First, the variance of the responses for the saliency algorithms (Fig 8C) is much lower than that of the human subjects (Fig 8A,B). Since the saliency algorithms are trained on human data, they highlight similar features in the images, particularly faces of humans and animals. Second, the percentage of correct responses monotonically improves as the target size, with respect to the parafovea, increases. Essentially, the larger the target, the higher are the chances that saliency maxima will fall within the ground truth mask. 

\section*{Acknowledgments}
The authors are grateful for the datasets and advice provided by Simon Thorpe, Molly Potter, and Brad Wyble. We thank Sang-Ah Yoo for assistance with designing the human experiment.

\nolinenumbers
%
%
%

\end{document}